\documentclass[10pt,twocolumn,letterpaper]{article}

\usepackage[pagenumbers]{cvpr} 

\usepackage{graphicx}
\usepackage{amsmath}
\usepackage{amssymb}
\usepackage{algpseudocode, algorithm}
\usepackage{booktabs}
\usepackage{multirow}

\usepackage[pagebackref,breaklinks,colorlinks]{hyperref}

\usepackage[capitalize]{cleveref}
\crefname{section}{Sec.}{Secs.}
\Crefname{section}{Section}{Sections}
\Crefname{table}{Table}{Tables}
\crefname{table}{Tab.}{Tabs.}

\def\paperID{0024} 
\def\confName{CVPR}
\def\confYear{2024}

\begin{document}

\title{How Much You Ate? Food Portion Estimation on Spoons}


\author{
Aaryam Sharma\textsuperscript{1}\thanks{Authors contributed equally.} , Chris Czarnecki\textsuperscript{1}\textsuperscript{*},
Yuhao Chen\textsuperscript{1}\thanks{Corresponding author.} , Pengcheng Xi\textsuperscript{2}, Linlin Xu\textsuperscript{1}, Alexander Wong\textsuperscript{1}\\
\textsuperscript{1} Vision and Image Processing Lab, University of Waterloo,
\textsuperscript{2} National Research Council Canada\\
{\tt\small a584sharma@uwaterloo.ca, cczarnec@uwaterloo.ca, yuhao.chen1@uwaterloo.ca} \\
{\tt \small pengcheng.xi@nrc-cnrc.gc.ca, l44xu@uwaterloo.ca, alexander.wong@uwaterloo.ca}
}

\maketitle

\begin{abstract}
Monitoring dietary intake is a crucial aspect of promoting healthy living. In recent years, advances in computer vision technology have facilitated dietary intake monitoring through the use of images and depth cameras. However, the current state-of-the-art image-based food portion estimation algorithms assume that users take images of their meals one or two times, which can be inconvenient and fail to capture food items that are not visible from a top-down perspective, such as ingredients submerged in a stew. To address these limitations, we introduce an innovative solution that utilizes stationary user-facing cameras to track food items on utensils, not requiring any change of camera perspective after installation. The shallow depth of utensils provides a more favorable angle for capturing food items, and tracking them on the utensil's surface offers a significantly more accurate estimation of dietary intake without the need for post-meal image capture. The system is reliable for estimation of nutritional content of liquid-solid heterogeneous mixtures such as soups and stews. Through a series of experiments, we demonstrate the exceptional potential of our method as a non-invasive, user-friendly, and highly accurate dietary intake monitoring tool.
\end{abstract}

\begin{figure*}
    \centering
    \includegraphics[width=0.95\textwidth]{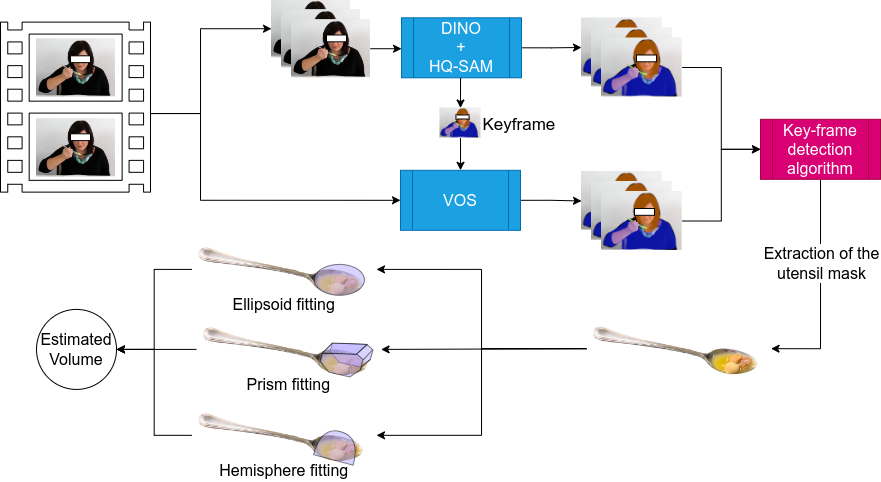}
    \caption{Summary of the pipeline. Videos are processed either by Grounding DINO and HQ-SAM or a VOS model (XMem). VOS requires single-frame priming, which is conducted using a single segmentation instance from HQ-SAM. Segmented frames are then filtered by Algorithm \ref{alg:ssf} and three different methods of volumetric estimation are evaluated. }
    \label{fig:pipeline}
\end{figure*}

\section{Introduction}
\label{sec:intro}

Most existing solutions in the market for nutritional intake estimation rely on self-reporting. However, nutritional intake data are often not entered immediately and the estimated values often have large errors that can reach up to $400\%$ when recalling intake after 24 hours \cite{binghamLimitationsVariousMethods2008}. Researchers have also found that self-reporting often leads to biased entries, where healthy foods are reported in larger amounts than unhealthy foods \cite{poslusna_misreporting_2009, kipnis_bias_2002}. It is also not be feasible for many elderly individuals or children to manually create a data entry in a tracking system before or after each meal \cite{cunhaEvaluationMSKinect2014a}.

Several alternative nutritional intake measurement approaches not involving self-reporting have been suggested over the years. Many approaches can be broken into three stages - food image classification and segmentation, food volume estimation, and nutritional content estimation \cite{hochsmann_review_2020, sultana_study_2023, konstantakopoulos_review_2024, subhi_vision-based_2019, resende_silva_survey_2017, wang_review_2022}. Among these stages, food volume estimation poses a particular challenge. It can be divided into 5 major types or approaches: 3D reconstruction, depth camera, multiple-view images, reference objects, model-based. \cite{konstantakopoulos_review_2024, sultana_study_2023}.

Estimation of nutritional content from photo or depth-camera captures of plates introduces two fundamental issues. Firstly, the depth of a dish might pose a problem for the camera to detect chunks of food in fluid media such as soups \cite{subhi_vision-based_2019}. Secondly, the individual consuming the food does not in all cases consume the entirety of the portion. Due to aging-related ailments this is especially common for elderly individuals \cite{kaiserFrequencyMalnutritionOlder2010}. Furthermore, certain food items such as soups are non-homogenous mixtures of solid objects occluded partially or fully by the liquid in which they are suspended. This creates a particular challenge for accurate volumetric estimation of each of the fractions in such a non-homogenous system.

We propose a method of nutritional intake tracking which revolutionizes the modality of tracking and addresses the designer biases encoded in currently available nutritional content estimation methods. The general outline of the pipeline we propose (Figure \ref{fig:pipeline}) is similar to most works that begin by using instance segmentation followed by food classification and volume recognition \cite{hochsmann_review_2020, sultana_study_2023, konstantakopoulos_review_2024, subhi_vision-based_2019, resende_silva_survey_2017, wang_review_2022}. We introduce an innovative solution that utilizes stationary user-facing cameras to track food items on utensils, not requiring any change of camera perspective after installation. The shallow depth of utensils provides a more favorable angle for capturing food items, and tracking them on the utensil's surface offers a significantly more accurate estimation of dietary intake without the need for post-meal image capture. The system is reliable for estimation of nutritional content of liquid-solid heterogeneous mixtures such as soups and stews. We test our approach on 10 short videos of moving rice on a spoon, the results of which we present in this paper, and obtain \textbf{21.9\%} MAPE (Mean Absolute Percentage Error) across the 10 videos on our best method. 

Our contributions include:
\begin{enumerate}
    \item A solution to the problem of estimating the volume of occluded food, by estimating the volume of food on utensils in videos.
    \item A model-based approach to perform the food volume estimation directly on the utensil used by the subject in a video capture.
    \item A novel algorithm to filter out spurious segmentation frames in the video capture of an individual consuming food.
\end{enumerate}


\section{Related Work}
Several methods have proven to be useful in the context of nutritional content and intake estimation. Reiterating the taxonomy defined by Sultana et al. \cite{sultana_study_2023}, the methods include 3D-reconstruction, depth-camera-based approaches, multi-angle camera view approaches, reference objects, model-based approaches and tracking from photo captures \cite{sultana_study_2023, hochsmann_review_2020, konstantakopoulos_review_2024, wang_review_2022}.

In 3D Reconstruction, various methods such as SLAM \cite{durrant-whyteSimultaneousLocalizationMapping2006,gaoFoodVolumeEstimation2018} are used to construct a 3D mesh of a captured food item from camera captures. Specifically, SLAM-based methods leverage visual odometry to approximate camera pose using keypoint features between multiple views and a number of corrective methods are administered to create a 3D representation of a scene \cite{gaoFoodVolumeEstimation2018}. Furthermore Gao \etal \cite{gaoFoodVolumeEstimation2018} use convex hull algorithm for surface reconstruction and propose an outlier-filtering algorithm which corrects the point-cloud representation.

Fundamentally, depth-based methods are similar to 3D reconstruction methods. However, no depth estimation needs to be performed, since the depth information is captured using a depth camera. A depth map is directly used to generate a point-cloud. Recently, deep learning approaches involving volumetric estimation using a point-cloud constructed with depth information have been introduced \cite{lo_point2volume_2020}. Recent works explore comprehensive 3D-model datasets of food items obtained using off-the-shelf hardware supporting direct depth capture and software facilitating construction of 3D meshes \cite{taiNutritionVerse3D3DFood2023}. Fewer methods focus on live tracking of the subject while they are consuming food.
The availability of RGBD cameras on the consumer market with devices such as XBox Kinect has prompted several researchers to explore the possibility of tracking nutritional intake from hand and face movements \cite{cunhaEvaluationMSKinect2014a}. Kassim \etal have developed a highly-accurate system for bite counting using a Kinect XBox Camera using hand and jaw movement tracking system \cite{kassimFoodIntakeGesture2019}. 

In methods based on multi-angle camera views, stereo-matching is first conducted to match features between multiple camera views. Camera position and angles are known and are used in the process. Subsequently, a point cloud is generated, which is then used to compute the volume by creating a mesh or using specific inference rules.

Reference objects are also used to address the problem of estimating real-world dimensions of the food item presented in the scene. Several kinds of reference objects have been used thus far, including the size of the bowl \cite{jia_estimating_2023} or the size of a thumb in a single photo \cite{pouladzadeh_measuring_2014}.

Lastly, some works use a model-based approach, where certain predefined 3D shapes are considered, such as a sphere or a cylinder that are subsequently fit to each detected food instance. Some methods propose fitting a wire-mesh model onto the food item \cite{sun_exploratory_2015}. This method simplifies the volume computation greatly as it eliminates costly calculation of volumetric approximations using integral calculus methods, often at the cost of some inductive bias incorporated into the estimation method. For example, \cite{sun_exploratory_2015} assumes a top-down view of the food item and does not perform automatic detection of the food type associated with the objects present on the plate.

Some methods also explore photo-capture-based approaches \cite{zhuTechnologyassistedDietaryAssessment2008,shao2021_ibdasystem,shao2021_nutri_database,he2021end,shao2023end}. Zhu \etal propose a data collection modality, where two photos are taken, one pre-meal and one post-meal, in order to estimate how much food has been consumed \cite{zhuTechnologyassistedDietaryAssessment2008}. Video-based tracking are superior to photo-based methods when addressing object occlusion due to the availability of a larger number of frames.

Several methods focus on video-based tracking. It is worth noting that prior works involving utensil tracking in particular originally focused on the idea of bite-counting, where a counter is increased upon the subject placing a portion of food in their mouth. Using RGB video capture Hossain \etal have developed a model for automated counting of bites and chews from a video capture \cite{hossainAutomaticCountBites2020a} leveraging affine optical flow to track the rotational movements of pixels in the regions of interests (ROI). However, works such as \cite{hossainAutomaticCountBites2020a} do not consider the fact that not all bites may equivalent. For example, one spoonful of soup can consist of only liquid or have one or more pieces of solid food of different nutritional density and different content present on it.

\section{Methodology}
\subsection{Overview}
\label{subsec:overview}

Our work focuses on videos and only analyzes the food on the utensil, rather than the entire container. It is probable that the video doesn't immediately depict the subject using a utensil with food on it, ready for consumption. Additionally, subjects often perform other tasks such as talking or watching a video during meals, resulting in an indeterminate gap between placing food on their utensil for volume estimation. This motivates the idea of filtering for frames in a video capture which do depict the subject holding a utensil with food on it prior to beginning our volume estimation. We thus add a key-frame detection step after the segmentation step, that serves as a start and end time for the volume estimation step.


For time efficiency, we select only one in 5 frames, as we find experimentally that this level of down-sampling does not negatively impact our results. 

Our pipeline depicted in Figure \ref{fig:pipeline} consists of three key steps: Zero-Shot Spoon Detection and Segmentation, Key-Frame Detection, and Volume Estimation.


\subsection{Zero-Shot Spoon Detection and Segmentation}
\label{subsec:segmentation}
One naive approach is to use zero-shot detection and segmentation models to segment utensils each frame. However, the resulting segmentations are inconsistent across consecutive frames (as shown in Figure \ref{fig:subject_with_fork} )
Instead, we segment and track.
The first step in our approach is to segment the video frames to find the food and utensil segmentation masks. 

In order to perform this step, we evaluated two approaches, the first using an instance segmentation model treating each frame independently, and the second being Video Object Segmentation. 

For the first approach, we use Segment Anything in High Quality (HQ-SAM) \cite{ke_segment_2023}. Due to the large variety of possible food types, it is an important consideration to choose a segmentation model that can segment several types of food. Furthermore, the size of the utensil and the food on it will usually constitute a small part of each frame in our videos. We therefore also require that a model that has a strong ability to detect well defined boundaries of object instances. HQ-SAM has been selected due to strong zero-shot performance and its sharpness of output segmentation masks. The benefit of such an approach is that it doesn't depend on the initial segmentation such as in Video Object Segmentation that could be of a really poor quality.

To obtain the instance segmentation from HQ-SAM, we prompt the model using bounding boxes obtained from Grounding DINO as it allows us to use text to prompt the model directly \cite{liu_grounding_2023} without the necessity of training a custom object detector. We use text-prompts in order to obtain bounding boxes for the objects in the scene in a zero-shot manner. These bounding boxes are subsequently fed-into HQ-SAM to obtain zero-shot segmentations. It further allows us to select the cutlery and any other types of instances that we require. For our model and the detection-and-tracking dataset, the prompts are "Food", "Face", "Spoon", "Fork". The usage of "Face" is important for the Key Frame detection algorithm we propose.

A combination of Grounding DINO and HQ-SAM is thus used to segment every selected frame. However, HQ-SAM is a large model with 2.45B total parameters which makes it resource-intensive. Furthermore, treating the frames independently doesn't use the temporal relation of adjacent frames. We also observe that this approach leads to several spurious segmentations. For example in several frames, the subject is classified as food along with the utensil.

To address these problems, we propose the usage of video object segmentation (VOS). Specifically, we propose XMem model \cite{cheng_xmem_2022}, as it is efficient, and is robust to occlusions \cite{cheng_xmem_2022}. In order to obtain the initial segmentation for VOS, we leverage HQ-SAM as before and we segment each frame. Once the key-frame detection algorithm identifies a key-frame, where the estimated volume is reasonable, we supply that frame to XMem, which is then used to segment subsequent frames in the capture. Not only is this approach faster, this leads to fewer spurious segmentations as well. We present Mean Intersection over Union (MIoU) scores for both approaches in Table \ref{tab:segmentation_results}.

\subsection{Key Frame Detection}
\label{subsec:key_frame}

Several challenges are inherent to the approach of volumetric estimation on utensils. Most videos have several frames containing utensils. However, only select keyframes can be used for food portion size estimation. Keyframes need to be selected according to the 6D-pose of the utensils and their position relative to the hands. Often occlusions can be observed, \eg when the food container occludes the utensil from the view, and instance confusion, \eg when the food is being picked up and the food instance on the utensil shares the same segmentation mask as the food in the container. Furthermore, there may be many utensils present in the view. 

\begin{figure}[htbp]
    \begin{subfigure}{\textwidth}
        \subfloat[\centering Bowl Segmented as food]{{\includegraphics[width=.15\textwidth]{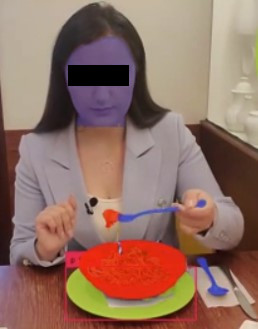} }}%
        \quad
        \subfloat[\centering Hand Segmented as food]{{\includegraphics[width=.146\textwidth]{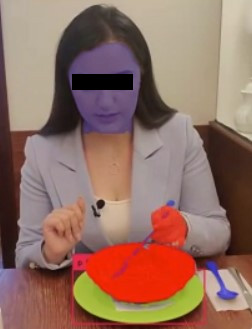} }}%
        \quad
        \subfloat[\centering Person Segmented As Food]{{\includegraphics[width=.144\textwidth]{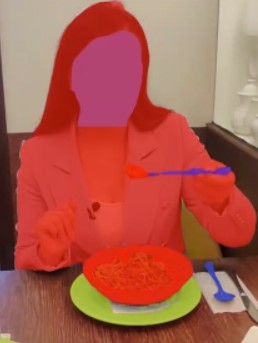} }}%
    \end{subfigure}
    \centering
    \caption{Examples of misclassified segmentation masks. In (a) the bowl has been improperly segmented as food. In (b) the hand has been improperly segmented as food. In (c) the independent frame segmentation method has mistakenly classifier the person as food!}
    \label{fig:subject_with_fork}
\end{figure}

Therefore, to estimate the volume accurately, we need to ensure that the food is on the fork or spoon and that the utensil is visible in the first place. To identify this instance in time, we select the spoon and fork instances with the highest confidence as per HQ-SAM's predictions. Then, for each segmented food instance thus obtained, we compute the distance of the food instance's centroid to the centroids of the highest-confidence spoon/fork instances, and pick the spoon/fork instance with the shortest distance. With the two pairs of (food, spoon) and (food, fork) instances, we select the pair that is closer to the individual's face. We provide an example in Figure \ref{fig:subject_with_fork}, where we see a woman holding a fork with spaghetti on it, and a spoon lying next to her. In this particular case, the fork with the spaghetti on it will be the pair considered for volume estimation, and the spoon will be ignored.
In order to identify the correct instance of cutlery holding the food, which is about to be eaten, we note also that in most scenarios, only one piece of cutlery is lifted up with food on it. This is also brought towards the mouth for consumption (Figure \ref{fig:subject_with_fork}). As a result, there is a point when the food on the spoon is sufficiently above the plate to avoid confusion between food instances. To quantify this point, we compute the distance between the centroid of the face of the person and the cutlery, and if it is below a certain threshold (currently 50\% of the image height for our dataset), then we compute the volume. Notably, this also improves the quality of the HQ-SAM segmentation in both approaches.

\subsection{Volume Estimation}
\label{subsec:volume}

Once utensil instance has been detected and segmented in the video, we can estimate the portion size of the food on the spoon based on their shapes observed in the key frames. We use three different 3D shapes: Prisms, Ellipsoids, and Hemispheres. We chose these shapes for their simplicity, given the limited general range of foods that can be fit on a utensil.

All of our methods use the mask of the food instance to determine the volume of the food rather than the mask or the length of the spoon. As a result, our volume estimation method dynamically adjusts to situations where the food on the utensil is of a different size or shape and fills a small portion of the spoon.

In all cases we assume width of the spoon as $3.81$ cm, which corresponds to the average table spoon width that we have obtained from physical measurements. 


\begin{figure*}[htbp]
    \centering
    \begin{subfigure}{\textwidth}
        \begin{center}
        \subfloat[\centering Prism Shape Fitting]        {{\includegraphics[width=.31\linewidth]{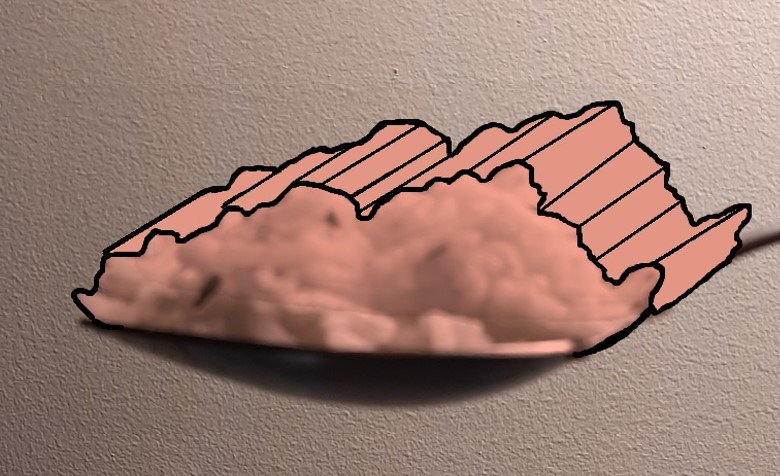} }}%
        \subfloat[\centering Hemisphere Shape Fitting]{{\includegraphics[width=.32\linewidth]{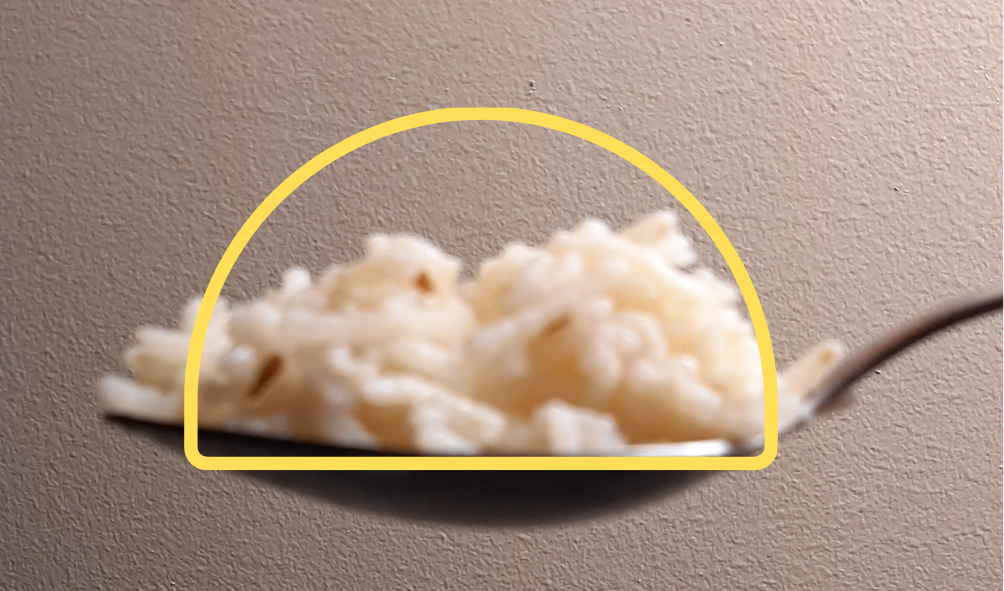} }}%
        \subfloat[\centering Ellipsoid Shape Fitting]{{\includegraphics[width=.32\linewidth]{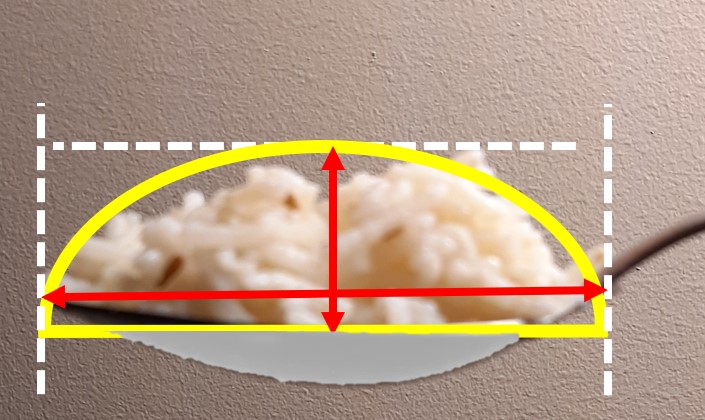} }}%
        \end{center}
    \end{subfigure}
    \caption{The different shape fittings. In (a) the prism shape that is fit based on the average depth set as an assumption $3.81$ cm for a spoon, the parallel sides of the prism being the food segmentation masks themselves. (b) shows the hemisphere fit based on the area. Both (a), (b) do not consider the curved bottom part of the spoon as the excess volume included in the fitted curves accounts for this. (c) shows the ellipsoid model where we compute the length and width of the segmentation to find the ellipsoid's axis lengths. We add the volume of the bottom part of the spoon which is assumed to be 5 $cm^3$}
    \label{fig:model_fitting}
\end{figure*}


\textbf{Approach 1: Prism Approximation}:
Treating the portion of food on the cutlery as a prism shape. If we estimate the per-pixel length for all pixels in the food instance, as well as estimate or measure the depth of the food portion instance, we compute the real area of the food instance and then multiply it with the width of the spoon. The prism shape is shown in Figure \ref{fig:model_fitting}a.  While a typical food instance on a spoon looks like an ellipsoid (Figure \ref{fig:model_fitting}b), this approach ignores the volume in the bowl of the spoon, which we factor as part of the excess volume predicted.

\textbf{Approach 2: Hemisphere Approximation}
Most food portions on a spoon or fork are not ideal prisms. In fact, on a spoon, they are usually irregular, but often can be approximated using a hemisphere. In this approach, we find the per-pixel area for the food instance. Approximating the instance as a hemisphere, we treat the portion visible as $\frac{1}{4}^{th}$ of the surface of a sphere, meaning that the area of the food instance mask is $\pi r^{2}$ and the volume of the food thus is $\frac{2}{3} \pi r^{3}$. Similarly to Prism Approximation, we ignore the volume of the spoon bowl, since the hemisphere method tends to capture excessive volumes.

\textbf{Approach 3: Ellipsoid Approximation}
Another method for approximating is to assume that the food on the spoon is an ellipsoid. We can then compute the volume of the ellipsoid by computing the length of the food instance, as well as it's height. For the spoon bowl volume, we take a constant surplus volume value of 5 $cm^3$ which we found by measuring the volume of water we can fit in an average table spoon. As shown in Figure \ref{fig:model_fitting}c, we see that an ellipsoid tends to closely fit to the food instance on the spoon, and doesn't have much excess volume, which necessitates the addition of the 5 $cm^3$ in volume to the final prediction.

In order to estimate the volume directly from the segmentation we must use some general facts about our utensils as well as the method with which they are used.

\begin{figure*}[htbp]
    \centering
    \begin{subfigure}{\textwidth}
        \begin{center}
        {{\includegraphics[width=.23\linewidth]{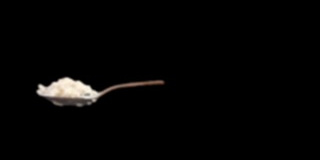} }}%
        {{\includegraphics[width=.23\linewidth]{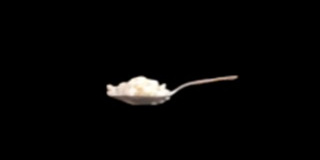} }}%
        {{\includegraphics[width=.23\linewidth]{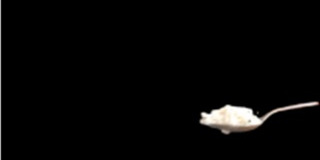} }}%
        {{\includegraphics[width=.23\linewidth]{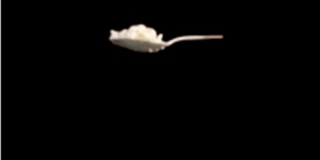} }}%
        \end{center}
    \end{subfigure}
    \caption{Representation of the spoon in different frames of a video capture of a subject eating. Captures similar to this have been used to test the utensil tracking and volumetric estimation accuracy of our proposed approach.}
    \label{fig:spoon_position}
\end{figure*}

We require a reference point in our image to estimate the dimensions of the food instance. To do this, we use the fact that most table spoons or forks have approximately the same size and dimensions for the same parts. Furthermore, in a video taken by a front-facing camera, people often hold the spoon/fork parallel to the camera frame (Figure \ref{fig:subject_with_fork}a,c, \ref{fig:spoon_position}). Although we assume spoon sizes to be consistent in world-dimensions, the per-pixel length of the spoon can be different in each frame due to changes in perspective and distance to the camera lens. We are therefore searching for a mapping from pixel-size to the real-world size of the spoon. Thus, we estimate the per-pixel length by identifying the endpoints of the bowl in a spoon-instance mask or the tip of the fork and the start of its neck, and we can use the average length of these parts of the cutlery to obtain a reference per-pixel length for the food, as the food is also in the same plane as the cutlery. Using a per-pixel mapping to the real length, we can work with foods that do not fully fill the spoon/fork as we can compute their dimensions independently of the portion present on the utensil.

To compute the per-pixel length of the food instance, using the total length of the spoon or fork is not ideal because the handle is often occluded by the subjects hand, which prevents us from getting a reliable utensil length in terms of pixels. We instead note that the part of the utensil on which the food is placed, i.e., the length of the bowl in a spoon, or correspondingly the distance of the point of the fork to its neck is fully visible in most frames. We thus seek to compute the length of this portion in pixels in order to find the desired per-pixel mapping to real length.

\begin{figure}
    \centering
    \rotatebox{0}{\includegraphics[width=0.8\linewidth]{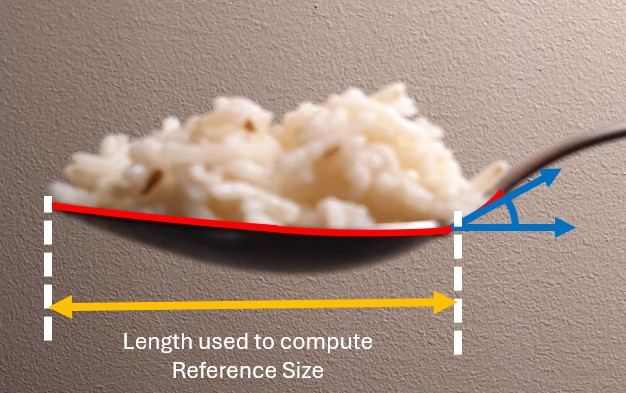}}
    \caption{Illustration of the reference length computation. We use the top curve of the spoon to find the first point from the tip of the spoon at which the slope of the curve is more than 15 degrees. In this diagram, the angle in blue is 30 degrees}
    \label{fig:reference_calculation}
\end{figure}

We make the further observation that in most spoons, there is a bend at the neck of the utensil. We leverage this abrupt change in angle to identify the bowl of a spoon or the portion between the point of the fork to its neck. We determined the angle of bend is 30 degrees for a spoon and 15 degrees for a fork. We found this by measuring the angle of the bend at the neck of a number of off-the-shelf spoons and forks which were readily available for purchase in local stores. We have found that the angle of the bend (blue angle in Figure \ref{fig:reference_calculation}) is approximately 30 degrees for a spoon and 15 degrees for a fork.

To compute the point of the bend, we fit a curve on the top surface of the utensil's segmentation mask and traverse the curve until we find the first point at which the gradient of the curve has an angle larger than 15 degrees for a spoon, or larger than 7 degrees for a fork. Once we find this x co-ordinate in our image, we use it along with the x-coordinate of the other end of the bowl of the spoon or tip of the fork to measure its pixel length, as shown in Figure \ref{fig:reference_calculation}. Note that we take lower thresholds of 15 and 7 degrees respectively instead of 30 and 15 degrees, in order to increase the chance of detection.

Furthermore, to reduce the effects of possible bumps in the mask, we take a 5-pixel moving average of the y-coordinate of the curve. The approximate gradient ($\nabla_c$) of this curve is computed between adjacent points with the formula:

\begin{equation}
    \nabla_c [i-1] = \frac{y[i] - y[i-1]}{x[i] - x[i-1]}
\end{equation}

where $[i]$ denotes the $i^{th}$ point coordinate on a discretized curve.


We go along the generated curve, and compute the angle between three points $i-1, i, i+1$ using the formula:
\begin{equation}
    \arctan(\nabla_c [i]) - \arctan(\nabla_c [i-1])
\end{equation}

We finally compute the per-pixel length using the average size of the bowl of a spoon or the distance between the tip and neck of a fork which is $6\ cm$ and $7.5\ cm$ respectively. These constants were also determined by measuring the same spoons and forks as for angle determination.

In this study, we focus on the feasibility of the method. We have chosen table spoons as the utensil for this study, since their shape is standardized and table spoons are widely used. We leave the analysis of different shapes and sizes of spoons to a future study. Only a slight modification to the assumed length and angle of the bend of the spoon and fork necks may need to be applied to work with other utensils.

\begin{table*}[]
\small
\centering
\begin{tabular}{c|c|c|cc|cc}
\hline
VOS & Shape & Alg. \ref{alg:ssf} & \multicolumn{1}{c|}{\begin{tabular}[c]{@{}c@{}}Per Frame \\ MAE (cm\textasciicircum{}3)\end{tabular}} & Per Frame MAPE & \multicolumn{1}{c|}{\begin{tabular}[c]{@{}c@{}}Final Prediction\\ MAE (cm\textasciicircum{}3)\end{tabular}} & \begin{tabular}[c]{@{}c@{}}Final Prediction\\ MAPE\end{tabular} \\ \hline
\multirow{6}{*}{\begin{tabular}[c]{@{}c@{}}Without \\ VOS\end{tabular}} & \multirow{2}{*}{\begin{tabular}[c]{@{}c@{}}Prism\\ Approx.\end{tabular}} & No Filter & 61.119 & 538\% & 51.780 & 457\% \\
 &  & Filter & 3.759 & 32.7\% & 2.748 & 23.4\% \\ \cline{2-7} 
 & \multirow{2}{*}{\begin{tabular}[c]{@{}c@{}}Hemisphere \\ Approx.\end{tabular}} & No Filter & 107.948 & 950\% & 93.910 & 831\% \\
 &  & Filter & 8.164 & 68\% & 8.164 & 68\% \\ \cline{2-7} 
 & \multirow{2}{*}{\textbf{\begin{tabular}[c]{@{}c@{}}Ellipsoid\\ Approx.\end{tabular}}} & No Filter & 26.697 & 233.6\% & 18.218 & 160.6\% \\
 &  & \textbf{Filter} & \textbf{3.444} & \textbf{29.8\%} & \textbf{2.675} & \textbf{21.9\%} \\ \hline
\multirow{6}{*}{\begin{tabular}[c]{@{}c@{}}Using VOS \\ (XMem)\end{tabular}} & \multirow{2}{*}{\begin{tabular}[c]{@{}c@{}}Prism \\ Approx.\end{tabular}} & No Filter & 4.395 & 36.7\% & 3.208 & 25.4\% \\
 &  & Filter & 3.803 & 31.7\% & 3.022 & 24.7\% \\ \cline{2-7} 
 & \multirow{2}{*}{\begin{tabular}[c]{@{}c@{}}Hemisphere \\ Approx.\end{tabular}} & No Filter & 9.312 & 77.9\% & 9.312 & 77.9\% \\
 &  & Filter & 9.115 & 76.2\% & 9.115 & 76.2\% \\ \cline{2-7} 
 & \multirow{2}{*}{\begin{tabular}[c]{@{}c@{}}Ellipsoid\\ Approx.\end{tabular}} & No Filter & 10.685 & 84.2\% & 7.207 & 58.5\% \\
 &  & Filter & 3.672 & 29.9\% & 3.245 & 26.1\% \\ \hline
\end{tabular}
\centering
\caption{The results of our experiments. The best predictions and lowest per-frame MAPE are obtained when an ellipsoid model with the Spurious Filtering algorithm and XMem is used. Notably, the spurious filtering algorithm is essential in obtaining this result, as the without this, there is a significantly larger prediction MAPE for the same geometric modelling method.}
\label{tab:ablation-results}
\end{table*}

\subsection{Post Processing}

Due to motion blur in certain frames, we notice that the segmentation results are often poor, which lead to poor volume computation. For example, we observe that occasionally the algorithm returns implausible food volumes, such as $100\ cm^3$ of food on a single spoon. We thus ignore the predictions for instances with volumes larger than 25 $cm^3$.



To further reduce the effect of these poor predictions due to motion blur, we employ the following algorithm:
\begin{algorithm}
\caption{Spurious Segmentation Filtering Algorithm}
\label{alg:ssf}
\begin{algorithmic}
    \State $ mean \gets 0$
    \State $ bad \gets 5$
    \For{\texttt{each frame sequentially}}
    \State $vol \gets \texttt{Volume of food on utensil in frame}$
    \If{$0 < vol < 25$}
        \State \texttt{Recompute the mean using vol and save it}
        \State $bad \gets 5$
    \Else
        \If{$bad = 0$}
            $mean \gets 0$
            \State \texttt{Store 0 as prediction}
        \Else
            \State \texttt{Store the previous mean as current volume}
            \State $bad \gets bad - 1$
        \EndIf
        \EndIf
    \EndFor
\end{algorithmic}
\end{algorithm}

In the Algorithm \ref{alg:ssf}, we maintain a running mean of good predictions, until we receive a consecutive sequence of $5$ bad predictions, after which we start recording bad predictions. This accounts for situations when the user is no long placing food on the utensil or when our segmentation step has not resulted in a good segmentation.

The final volume estimation is computed by taking the average of all the measurements taken after the key-frame detection. An average of the predictions for the output reduces the impact of noise in our predictions and uses the maximum usable frames in our video. Furthermore, it reduces our reliance on a single well-timed segmentation that is difficult to obtain in the entire frame sequence.

\section{Data Collection}

For the purposes of testing this approach, we filmed 10 videos 8 - 12 seconds in length. We placed rice on a spoon from a small container, and we moved the spoon with rice on it in diverse directions. To measure the volume of rice on the spoon, the water displacement method was used, by placing the rice in a container of water and then measuring the volume of water it displaced.

The aforementioned 10 videos are shot in the same setting, under the same conditions and contain the same subject operating the utensil. Some of the positions can be seen in Figure \ref{fig:spoon_position} with the background and subject removed.

We also collected a dataset to test out the instance-segmentation ability of our dataset, where we computed the Mean Intersection over Union for the correct spoon and food instances. This dataset consists of frames taken at intervals of 20 frames from 5 different YouTube videos.


\section{Results}

Segmentation MIoU results have been summarized in Table \ref{tab:segmentation_results}.

There are 12 volume estimation approaches tested, outlined in Table \ref{tab:ablation-results}. Ablations were conducted for the end-to-end pipeline covering different segmentation, volumetric approximation methods and the usage of Algorithm \ref{alg:ssf} or lack thereof.

Each aforementioned approach has been tested on 10 videos, which the authors have recorded. Volume of each spoonful has been determined by means of water displacement. We compute the Mean Absolute Error (MAE) between the actual measured volume of rice as compared to the final predicted volume for each video. 

Notably, the approximate value of volume of food on a spoon is in the range of 10 - 17 $cm^3$ in our dataset. We leverage the Mean Absolute Percentage Error metric (MAPE) to account for the variable volume of food placed on an utensil in each video sample. Illustratively, an error of 5 $cm^3$ is relatively small in a video where there is 17 $cm^3$ of rice on the spoon as compared to a video where there is 10 $cm^3$ of rice.

We also report the Mean Absolute Error (MAE) between the actual volume and the predicted volume for each frame. The per-frame MAE is computed for each video, and then the resulting MAEs are averaged. This is done to prevent metric skew which would be associated with videos of longer duration. The motivation for computing the per-frame error is to illustrate the noise in the predictions for each frame.

Table \ref{tab:ablation-results} indicates that the approaches which use the Algorithm \ref{alg:ssf} are vastly superior as compared to the approaches without the Algorithm \ref{alg:ssf} when we do not use VOS, as we find the error for those approaches to be more than three times the maximum expected volume of rice on a spoon. The best-performing method was Ellipsoid approximation with Algorithm \ref{alg:ssf} using independent frames segmentation.

The main reason for abnormally large volume predictions is the incorrect segmentation of food items. In several frames, it can be seen that occasionally a subject is misclassified as a food instance (see Figure \ref{fig:subject_with_fork}c).

However, when we use the segmentations from XMem, such misclassifications are less frequent. Thus the result without Algorithm \ref{alg:ssf} does not have as high of an error discrepancy from when the Algorithm \ref{alg:ssf} is used. The spurious segmentation filtering algorithm exhibits a marginal impact on the per-frame MAPE for Prism and Hemisphere VOS based approximations. However, for the ellipsoid approximation method, Algorithm \ref{alg:ssf} is essential even in the VOS case. This is because, the computation of the axes of the ellipsoid requires us to find the left, right, top and bottom endpoints of the food instance's segmentation. If the segmentation has spurious protrusions in any direction, this can lead to at least one of the axes to be improperly large.

\begin{table}[]
\small
\begin{tabular}{c|c|c}
\hline
Estimation Method  & Spoon MIoU & Food MIoU \\ \hline
Independent Frames & 0.437      & 0.0896    \\
VOS                & 0.139      & 0.0138    \\ \hline
\end{tabular}
\centering
\caption{The results of both segmentation approaches on our segmentation dataset}
\label{tab:segmentation_results}
\end{table}

On our segmentation dataset, we find that using independent frames without VOS outperforms VOS greatly as seen in Table \ref{tab:segmentation_results}. We note that VOS performs poorly here because the frames selected are far apart (20 frames), and thus VOS is unable to make use of temporal features effectively. Another challenge for the models in this dataset for food instance segmentation was that in the YouTube videos the subjects face is very close to the plate, and hence the food on the spoon can often get confused with the food on the plate. As a result, the MIoU for the food on spoon instance is quite low.

\section{Discussion}

Algorithm \ref{alg:ssf} eliminates frames with entire objects being labelled incorrectly, whereas for the frames where the instances is correct, HQ-SAM allows us to obtain crisper and cleaner segmentation masks, thus leading to better volume predictions.


It is worth pointing out that the error in the final prediction of the hemisphere approach and the per-frame errors are identical for the VOS variant because the volume predictions produced by this approach are consistently smaller than the ground-truth value. We attribute this to underestimating the volume of the food on the bowl of the spoon when the portion is not visible in the frame.

For proceeding in the wild, we believe that the Prism approach will be the most reliable method among the described methods on more diverse food types. For example, noodles on a fork do not conform to neither the ellipsoid model nor the hemisphere model. However, a prism shape can be fit more tightly around the food instance. This also inspires approaches that combine multiple geometric shapes for each instance on the utensil, that are chosen once the food instance is classified or its shape is analyzed.

Noting the results on our segmentation dataset, we further emphasize the use of Spurious Segmentation Filtering, i.e., Algorithm \ref{alg:ssf}, since we find that there are numerous frames with poor segmentation of the food on the spoon, and without filtering these, we are unable to obtain a reasonable volume estimation. We thus also recommend users not to introduce large gaps between frames in order to improve efficiency, as it reduces the performance of the VOS method greatly.

\begin{table}[]
\small
\begin{tabular}{c|c|c|cc}
\hline
VOS & Shape & Filter & \multicolumn{1}{c|}{\begin{tabular}[c]{@{}c@{}}Best Frame \\ MAE (cm\textasciicircum{}3)\end{tabular}} & \begin{tabular}[c]{@{}c@{}}Best Frame \\ MAPE\end{tabular} \\ \hline
\multirow{6}{*}{\begin{tabular}[c]{@{}c@{}}W/o\\ VOS\end{tabular}} & \multirow{2}{*}{\textbf{Prism}} & \textbf{No} & \textbf{0.534} & 4.45\% \\
 &  & Filter & 1.145 & 9.36\% \\ \cline{2-5} 
 & \multirow{2}{*}{\begin{tabular}[c]{@{}c@{}}Hemi-\\ sphere\end{tabular}} & No & 2.241 & 22.59\% \\
 &  & Filter & 5.328 & 43.97\% \\ \cline{2-5} 
 & \multirow{2}{*}{\begin{tabular}[c]{@{}c@{}}Ellip-\\ soid\end{tabular}} & No & 0.629 & 5.46\% \\
 &  & Filter & 1.418 & 11.8\% \\ \hline
\multirow{6}{*}{\begin{tabular}[c]{@{}c@{}}Using \\ XMem\end{tabular}} & \multirow{2}{*}{\textbf{Prism}} & \textbf{No} & 0.577 & \textbf{4.43\%} \\
 &  & Filter & 1.456 & 11.88\% \\ \cline{2-5} 
 & \multirow{2}{*}{\begin{tabular}[c]{@{}c@{}}Hemi-\\ sphere\end{tabular}} & No & 7.348 & 60.91\% \\
 &  & Filter & 8.025 & 67.10\% \\ \cline{2-5} 
 & \multirow{2}{*}{\begin{tabular}[c]{@{}c@{}}Ellip-\\ soid\end{tabular}} & No & 1.479 & 11.1 \\
 &  & Filter & 2.339 & 18.8 \\ \hline
\end{tabular}
\centering
\caption{The best frame prediction. This table shows the MAE and MAPE of the best frame that was predicted by each method in each video. We can see that the Prism approach has the best frame MAE and MAPE without Algorithm \ref{alg:ssf}
}
\label{tab:best_frame_prediction}
\end{table}


\section{Conclusion}

In this paper we have presented a novel approach towards nutritional intake tracking, which addresses the unique challenges posed by video and photo tracking modalities. We have described a multi-stage end-to-end volumetric estimation pipeline consisting leveraging modern image and video segmentation models and proposed an on-utensil tracking approach. We hope our contribution will lead to considerable progress towards the development of more accessible and user-friendly nutritional content tracking solutions.

\section{Acknowledgement}
This work was supported by the National Research Council Canada (NRC) through the Aging in Place (AiP) Challenge Program, project number AiP-006. The authors also thank Qasim Ali and Nixon Chan for their assistance with data collection and annotation.

{\small
\bibliographystyle{ieee_fullname}
\bibliography{references}
}

\end{document}